\documentclass[letterpaper, 10 pt, conference]{ieeeconf}  %

\usepackage{array}
\usepackage{stackengine}
\usepackage{xspace}
\usepackage{graphicx}
\usepackage{subcaption}
\newcommand{\grenderwidth}{0.19\textwidth}
\usepackage{caption}
\usepackage{xcolor}
\usepackage{amssymb, amsmath, mathtools}
\usepackage{graphicx}
\usepackage{float}
\usepackage{multirow}
\usepackage{booktabs}
\usepackage{svg}
\usepackage{subcaption}
\usepackage[utf8]{inputenc}
\usepackage[T1]{fontenc}
\usepackage[english]{babel}
\usepackage{svg}
\usepackage{overpic}
\usepackage{algorithm}
\usepackage{algpseudocode}
\usepackage[colorlinks, citecolor=gold, linkcolor=gold, urlcolor=gold]{hyperref}
\usepackage[nameinlink,capitalise]{cleveref}
\makeatletter
\let\NAT@parse\undefined
\makeatother
\usepackage[numbers,sort&compress]{natbib}

\definecolor{gold}{HTML}{daaa00}

\DeclareMathOperator*{\argmin}{arg\,min}

\IEEEoverridecommandlockouts                              %

\title{\LARGE \bf
TGrasp6D: Task-Informed Vision-Language Priors and Pose Refinement enable Zero-Shot Task-Oriented 6DoF Grasping
}

\author{Manav Kulshrestha$^{1}$, Apoorva Vashisth$^{1}$, Damon Conover$^{2}$, Aniket Bera$^{1}$%
\thanks{\raggedright $^{1}$Manav Kulshrestha, S. Apoorva Vashisth, and Aniket Bera are with the IDEAS Lab, Department of Computer Science, Purdue University, West Lafayette, IN, USA. %
        {\tt\small \{\href{mailto:mkulshre@purdue.edu}{mkulshre}\allowbreak,\href{mailto:vashista@purdue.edu}{vashista}\allowbreak,\href{mailto:aniketbera@purdue.edu}{aniketbera}\allowbreak\}@\allowbreak purdue\allowbreak.edu}}
\thanks{\raggedright $^{2}$Damon Conover is with the DEVCOM Army Research Laboratory, Adelphi, MD, USA. %
          {\tt\small \href{mailto:damon.m.conover.civ@army.mil}{damon.m.conover.civ@army.mil}}}%
}

\begin{document}

\maketitle
\thispagestyle{empty}
\pagestyle{empty}

\begin{abstract}
Task-Oriented Grasping (TOG) has emerged as an important skill for purposeful robotic manipulation. 
Given the task of using a knife to ``slice bread,'' a robot must choose to grasp from the handle, whereas for ``hand[ing] the knife over,'' it is preferable for the handle to remain unobstructed for safe handoff. 
Current approaches for TOG either focus on 2D planar grasping, remain limited by their reliance on task-agnostic grasp generators, or are constrained by reliance on finite, manually-annotated, task-specific grasp datasets that cannot cover all possible grasp-task pairs. %
To that end, we present TGrasp6D, a vision-language based approach for generating task-oriented 6DoF grasps in a zero-shot manner. 
Given a task and a single RGB-D image of an associated object, TGrasp6D reasons about the appropriate semantic affordance for the given task and object geometry to produce an initial task-informed prior. 
This generated prior is further refined via 6DoF pose optimization for collision avoidance while preserving task consistency. 
Our approach achieves state-of-the-art success rate of 94\% on the TaskGrasp dataset, with the best baseline at 74\%. 
We also deploy our approach on a Franka Research 3 robot to demonstrate zero-shot generalization to real world objects.
\end{abstract}

\section{Introduction}

Robotic manipulation, and specifically robotic grasping, is a fundamental capability for enabling robots to physically interact with, and operate, in the real world~\cite{vashisth2026corelin, kulshrestha2023structural}. 
Recent methods have significantly advanced this field~\cite{kulshrestha2025vlad, bukhari2025variational, iwase2025zerograsp} by prediction of stable grasps for diverse objects. 
However, a stable grasp solves half the problem: it does not consider whether the resulting hand-object configuration will support the intended task. 
Task-Oriented Grasping (TOG) captures this coupling between grasp and intent, requiring joint reasoning over object affordances and task semantics, rather than treating grasp selection as a purely geometric problem. 
This matters wherever robots operate around humans or chain manipulation steps together -- an assistive robot handing over a tool, or an industrial arm that must re-grasp a part differently depending on whether it's inserting or fastening.

\begin{figure}[htbp]
    \centering
    \includegraphics[width=\linewidth]{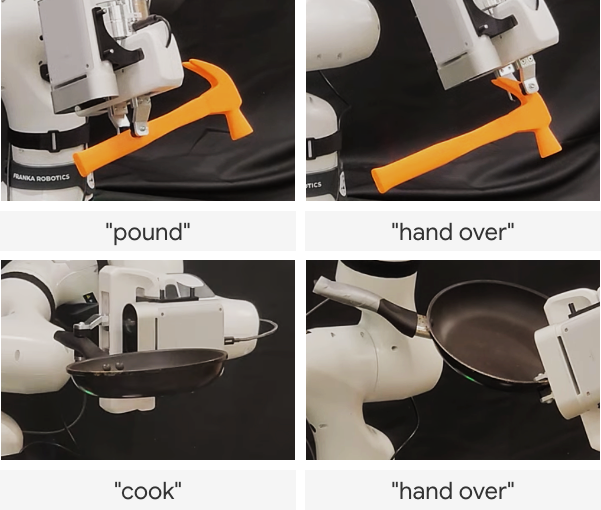}
    
    \caption{Example of TGrasp6D producing distinct 6DoF grasps for different tasks on the same real-world object. ``handover'' task results in grasps that keep the handle clear whereas tasks like ``pound'' and ``cook'' that require holding the hammer and pan from the handle are reflected.}
    \label{fig:placeholder}
    \vspace{-6mm}
\end{figure}

Various methods have been proposed for TOG. 
Training-based approaches~\cite{tang2023graspgpt, tang2025foundationgrasp, deshpande2025graspmolmo, murali2021same} rely on large human-annotated datasets of object-task-grasps labels~\cite{murali2021same}.
However, due to the boundedness of these datasets in the space of valid grasps and possible tasks, even large annotation efforts~\cite{deshpande2025graspmolmo} yield sparse coverage.
Hence, their performance~\cite{tang2023graspgpt, tang2025foundationgrasp, deshpande2025graspmolmo, murali2021same} is capped by the dataset size, and generalization to novel task-object combinations remains limited. 
Many approaches~\cite{tang2023graspgpt, tang2025foundationgrasp, dong2025rtagrasp, huang2025hgdiffuser, tang2025affordgrasp} attempt to bypass the generative aspect of grasp synthesis by coupling with a task-free generator~\cite{sundermeyer2021contact, yuan2023m2t2, fang2023anygrasp} and then scoring the resulting grasps for the given task, essentially reducing the problem to that of score regression.
By injecting task semantics onto task-agnostic grasp priors, these methods are bottlenecked by the coverage of the upstream generator: if no candidate near the task-relevant region was proposed, no downstream semantic reasoning can recover it.
Finally, planar-grasp methods~\cite{vuong2024language, holomjova2023one, tang2023task, li2024shapegrasp} restrict prediction to the image plane, discarding a broad class of feasible grasps with out-of-plane approach directions.%

In contrast, large vision-language models offer a different foundation for TOG.
Pretrained on internet scale image-text corpora, they encode priors over object semantics and functional affordances without requiring the exhaustive supervision that prior TOG methods depend on.
These priors, however, are expressed in the image and text modalities that the models were trained on.
Extending their reasoning requires situating the model in the three-dimensional, geometric space where manipulation actually occurs.

Hence, we present TGrasp6D, a framework that leverages VLMs to synthesize 6DoF task-oriented grasps in a zero-shot manner, requiring no task-specific training or fine-tuning.
Given a single RGB-D observation of an object and a natural language task instruction, TGrasp6D first reasons jointly over the object's geometry and semantics to identify the task-relevant grasp affordance, then localizes this affordance within the image.
To bridge the resulting image-space localization into 3D, we lift it into a plane consistent with the camera's projective geometry, yielding a task-informed grasp prior around the localized affordance.
Candidate grasps sampled within this prior are then refined through pose optimization that trades off collision avoidance against consistency with the task-relevant region, yielding grasps that are simultaneously geometrically feasible and appropriate to the task at hand. 
In summary, we make the following four contributions:
\begin{itemize}
\item We introduce \textbf{TGrasp6D}, a zero-shot framework for task-oriented 6DoF grasping from a single RGB-D observation and natural-language task instruction.

\item Our VLM-guided grasp synthesis pipeline grounds task-relevant affordances in image space, lifting them into 3D task-informed grasp priors, and refining candidate grasps through collision-aware pose optimization.

\item TGrasp6D achieves a \textbf{94\% success rate} on the TaskGrasp benchmark, outperforming the strongest baseline by $\mathbf{\sim20\%}$.

\item We demonstrate zero-shot transfer to real-world manipulation on a Franka Research 3 robot across novel objects and task instructions.

\end{itemize}

\section{Related Work}

\subsection{Task-Agnostic 6DoF Grasp Synthesis}

Early methods~\cite{sundermeyer2021contact, fang2023anygrasp, yuan2023m2t2, iwase2025zerograsp, bukhari2025variational, kulshrestha2025vlad} focused on generating stable 6DoF grasps conditioned only on object geometry, regardless of the downstream task. Sundermeyer et al. \cite{sundermeyer2021contact} root grasp contact points directly in the observed point cloud to reduce the search space for cluttered-scene grasping. Fang et al. \cite{fang2023anygrasp} learn dense, temporally-consistent grasp perception from large-scale supervision, incorporating object center-of-mass to improve stability. Yuan et al. \cite{yuan2023m2t2} unify grasping and placement in a single transformer that can optionally condition on a language goal, though this selects which object to act on rather than where on it to grasp. Iwase et al. \cite{iwase2025zerograsp} instead couple grasp prediction with 3D shape reconstruction, arguing that reasoning jointly about occluded geometry benefits both. Bukhari et al. \cite{bukhari2025variational} learn a probabilistic latent space over object shapes and use the inferred features to guide $\mathbb{SE}(3)$ diffusion-based grasp synthesis, improving robustness to noisy or partial observations over deterministic shape encoders. Most recently, Kulshrestha et al. \cite{kulshrestha2025vlad} removes the need for grasp-specific training by querying a VLM to generate a candidate antipodal axis in image space, then lifting it into 3D via depth alignment. While these methods differ in representation and supervision, all optimize purely for stability and feasibility; we instead condition the search for a feasible pose on a task-relevant region identified before optimization begins.

\subsection{Task-Oriented Grasping}

A first line of work learns task-conditioned grasp selection from annotated data. Murali et al. \cite{murali2021same} introduce TaskGrasp, the largest TOG benchmark to date, and propagate a semantic knowledge graph over objects and tasks to generalize to held-out instances. Tang et al. \cite{tang2023graspgpt} replace this closed-world graph with open-ended knowledge queried from an LLM to generalize to novel concepts, and \cite{tang2025foundationgrasp} extends the idea to both LLMs and VLMs over an expanded dataset. Deshpande et al. \cite{deshpande2025graspmolmo} scale data rather than knowledge sources, fine-tuning a VLM on a large synthetic corpus of cluttered, realistically-instructed scenes. Chu et al.~\cite{chu2025graspcot} instead incorporate explicit physical-property reasoning over multi-step, flexible language instructions to condition 6DoF grasp generation. Since each method depends on a bespoke annotated or synthesized TOG corpus, generalization is bounded by that corpus's coverage of the object-task-grasp space.

A second line of work couples task reasoning with an existing task-agnostic generator, applying semantics only after a candidate set has been produced. Dong et al. \cite{dong2025rtagrasp} retrieve grasp constraints from human demonstration videos and use them to filter task-agnostic candidates. Huang et al. \cite{huang2025hgdiffuser} fold similar human-derived constraints directly into a guided diffusion sampler rather than filtering after the fact, while Tang et al. \cite{tang2025affordgrasp} use a VLM to ground task-relevant affordance regions but still select among poses returned by a fixed generator \cite{fang2023anygrasp}. Because semantics are applied downstream of an already-fixed candidate distribution, a task-appropriate grasp the generator never proposed cannot be recovered by any amount of subsequent reasoning.

A third line of work restricts prediction to the image plane. Vuong et al. \cite{vuong2024language} treat grasp detection as a diffusion-based generative problem conditioned on language, Holomjova et al. \cite{holomjova2023one} predict a task-relevant grasp rectangle via one-shot learning, Tang et al. \cite{tang2023task} ground task and object jointly through a 2D vision-language model restricted to top-down grasping, and Li et al. \cite{li2024shapegrasp} decompose objects into convex parts for LLM-guided part selection but likewise execute largely top-down grasps. A task-relevant affordance need not lie in the image plane, so planar prediction structurally excludes a broad class of otherwise feasible grasps. TGrasp6D avoids all three limitations, requiring no task-specific dataset, no upstream task-agnostic generator, and no restriction to the image plane.

\begin{figure*}[htbp]
    \vspace{2mm}
    \centering
    \includegraphics[width=0.97\linewidth]{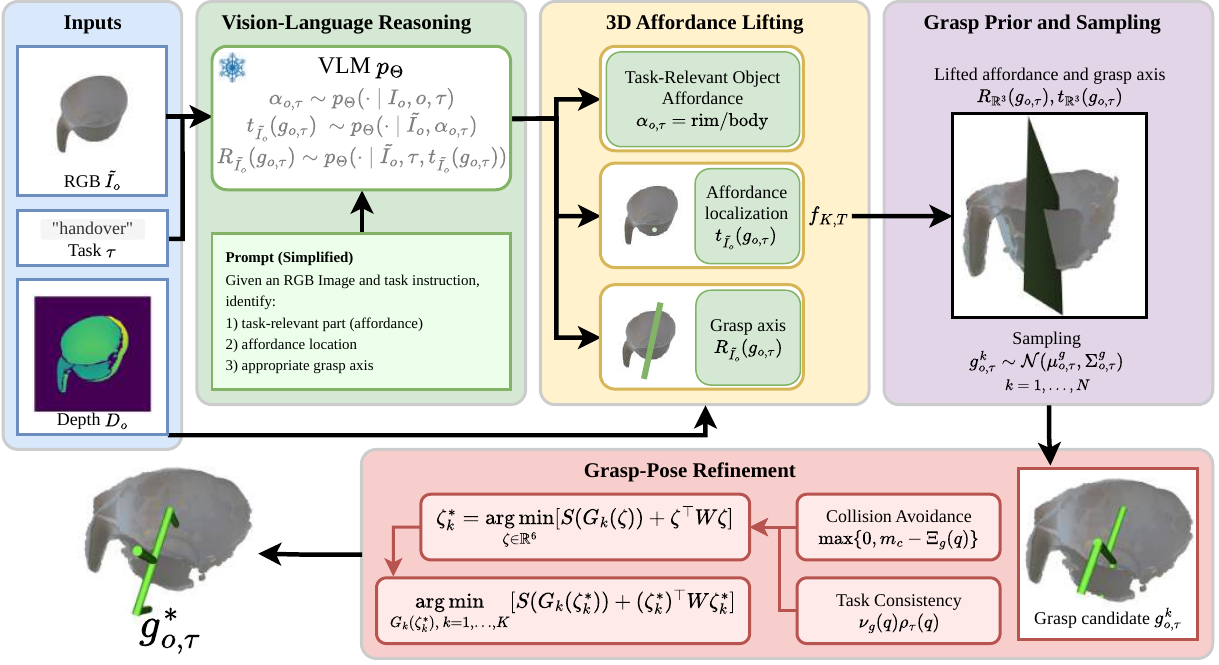}
    \caption{ Overview of TGrasp6D. Given a single RGB-D observation and a natural-language task instruction, a vision-language model identifies and localizes the task-relevant affordance and predicts an image-space grasp orientation. These cues are lifted into 3D to form a task-informed grasp prior, from which candidate 6DoF grasps are sampled and refined for collision avoidance while preserving task consistency, yielding the final task-oriented grasp.} 
    \label{fig:approach}
    \vspace{-5mm}
\end{figure*}

\subsection{Vision-Language Models for Affordance Reasoning in Manipulation}

Beyond grasping, VLMs increasingly ground task-level affordances for general manipulation. Huang et al. \cite{huang2023voxposer} show an LLM can infer free-form affordances and, by querying a VLM, compose them into 3D value maps that a planner optimizes over, demonstrating that language-model priors transfer zero-shot to full trajectory synthesis. Yuan et al. \cite{yuan2024robopoint} instruction-tune a VLM to output image keypoint affordances directly, generalizing across navigation, manipulation, and augmented reality, though the output remains a 2D point requiring a separate module to become an executable action. Qian et al. \cite{qian2024thinkgrasp} use a VLM's contextual reasoning to decide which occluding objects to remove in clutter, reasoning at the level of object visibility rather than part-level grasp selection on the target itself. \cite{tang2025affordgrasp} and \cite{deshpande2025graspmolmo}, discussed above, are the closest prior instances of VLM reasoning applied specifically to grasp affordances, but the former delegates pose synthesis to a task-free generator and the latter requires fine-tuning on a large synthetic corpus. We instead use a VLM purely to localize the task-relevant affordance, then lift and refine that region directly into a feasible 6DoF grasp through geometry-aware pose optimization.

\section{Problem Formulation}

Let $o \in \mathcal O$ be an object, represented by an image $I_o \in \mathcal I$ and depth map $D_o \in \mathcal D$ obtained by an observation function $\Omega: \mathcal O \to \mathcal I \times \mathcal D$. 
Consider $\tau \in \mathcal T$ to be an associated task instruction.
Furthermore, let $\psi: \mathbb{SE}(3)\times \mathcal O \to \{0, 1\}$ be a predicate which is satisfied, $\psi(g, o)$, when the grasp $g$ is stable for object $o$, and let $\varphi: \mathbb{SE}(3)\times\mathcal O\times\mathcal T \to \{0, 1\}$ be a predicate which is satisfied $\varphi(g, o, \tau)$ when the grasp $g$ is appropriate for the object $o$ and task $\tau$.

We define the problem of 6 degree-of-freedom task-oriented grasp synthesis from a single observation of the object, $<I_o, D_o>$, as producing a grasp $g_{o,\tau} \sim \pi_\theta(\cdot\ |\ I_o, D_o, \tau)$, which satisfies $\psi(g_{o,\tau}, o)\land\varphi(g_{o,\tau}, o, \tau)$ where $(I_o, D_o) = \Omega(o)$, and $\pi_\theta: \mathcal I\times\mathcal D\times\mathcal T \to \mathbb{SE}(3)$ is the synthesis policy, parametrized by $\theta$. 
We also define $t_{\mathcal X}: \mathbb{SE}(3) \to \mathcal X$, which maps a grasp $g \in \mathbb{SE}(3)$ to its translation $t_{\mathcal X}(g)$ and $R_{\mathcal X}: \mathbb{SE}(3) \to \mathcal X$, which maps the grasp $g$ to its rotation $R(g)$ in $\mathcal X$. 
For our purposes, $\mathcal X = \{\mathbb R^3, \mathbb{SO}(3), I\}$ where $I \in \mathcal I$ is an image.

\section{Methodology}

Our approach, TGrasp6D, consists of 3 major components: (1) VLM based object-centric task-oriented reasoning and localization of the appropriate object part affordance, (2) generation of a vision-language-based grasp-pose prior around the localized object part in image space and utilization of camera extrinsics for lifting it into 3D, and (3) sampling of candidate grasps from the grasp-pose prior and batched 6DoF pose refinement for collision avoidance and task-region consistency via optimization. The overview of our approach is detailed in Fig.~\ref{fig:approach}.

\subsection{Task-Oriented Reasoning}
Broadly, language models are trained on data $X$ and learn an implicit joint prior distribution $p_\Theta(\mathcal C)$ over the space of concepts $\mathcal C$ represented in the data $X$.
At inference, they can autoregressively sample from this prior to generate outputs $s \sim p_\theta(\cdot\ |\ \mathcal S)$, conditioning on any subset $\mathcal S \subseteq C$.
By definition, $\mathcal{S}$ is also represented in the data $X$.
For our purposes, it is sufficient to consider $\mathcal S = \mathcal O \cup \mathcal T$ as the joint space of all objects $\mathcal O$ and tasks $\mathcal T$.
State-of-the-art VLMs are trained on internet-scale data $X$ which ensures $\mathcal S = \mathcal O \cup \mathcal T \subseteq \mathcal C$, allowing us to leverage these models for our aim of generating task-oriented grasps for objects.

Our approach reasons about the object $o$ geometry in context of the task instruction $\tau$. 
We obtain a natural language representation of the task-relevant object affordance
\begin{equation}
\alpha_{o,\tau} \sim p_\Theta(\cdot\ |\ I_o, o, \tau)    
\end{equation}
where $I_o$ is the image representation of the object.
For an object $o=$``sword'' and task $\tau_s=$``stab,'' the associated task-relevant object affordance $\alpha_{o,\tau_s}$ may be ``hilt,'' whereas for a different task $\tau_h=$``hand over'' for the same object, the task-relevant object affordance $\alpha_{o,\tau_h}$ may be ``blade.'' (To safely hand the object to the human user).
The next step is to obtain a localization $t_{I_o}(g_{o,\tau})$ of the task-relevant object affordance $\alpha_{o,\tau}$.
Utilizing a VLM for processing this can be tricky since VLMs are prone to spurious distractors~\cite{sharma2024losing} and often perform poorly in conventional object-detection tasks without specific fine-tuning~\cite{zhong2022regionclip}.

To minimize these distractions, we utilize an object mask $M_o$ to exclude the object $o$'s background. 
We additionally perform a close cropping of the image so the object dominates the image foreground, producing $\tilde{I_o}, r_o = F(I_o, M_o)$. 
The overall destructive image transformation $F: \mathcal I \to \mathcal I\times\mathcal R$ additionally produces the residual information $r_o \in \mathcal R$, needed for inverting $F$, so that we can later recover the final generated grasp in the original image space $I_o$.
For obtaining the mask, we utilize an off-the-shelf image segmentation model~\cite{carion2026sam}.
With this, our approach localizes 
\begin{equation}
t_{\tilde{I_o}}(g_{o,\tau}) ~\sim p_\Theta(\cdot\ |\ \tilde{I_o}, \alpha_{o,\tau})
\end{equation}
the task-relevant object affordance $\alpha_{o,\tau}$ in image space.
In practice, the task-relevant object affordance localization $t_{\tilde{I_o}}(g_{o,\tau}) \in \mathbb R^2$ is represented as a 2D point indicating the location of the task-relevant object affordance $\alpha_{o,\tau}$ on the image $\tilde{I_o}$ of object $o$.

\subsection{Grasp Prior Generation}

Having obtained a task relevant object affordance $\alpha_{o,\tau}$, TGrasp6D has deduced \textit{where} to grasp the object $o$.
However, depending on the object $o$ geometry and task $\tau$ semantics, this is not sufficient to generate either a stable $\psi(g, o)$ or task-appropriate $\varphi(g, o, \tau)$ grasp in the image plane, much less in 3D space.
While lifting the localization $t_{\tilde{I_o}}(g_{o,\tau})$ to a single point in 3D will provide us with the translation of $g_{o,\tau}$, we still need to produce the rotation $R_{\mathbb{SO}(3)}(g_{o,\tau})$ to recover the full grasp $g_{o,\tau} \in \mathbb{SE}(3)$.
As a motivating example, consider grasping an object $o=$``mug'' for the task of $\tau_d=$``drinking.'' Identifying a possible affordance in $\alpha_{o,\tau_d}=$ ``handle'' (and localizing it) still requires the reconciliation between the grasp and the object such that the closing axis of the gripper must be perpendicular to the length of the mug's handle.

Hence, after generating task relevant object affordance, our approach generates the grasp orientation in image space -
\begin{equation}
    R_{\tilde{I_{o}}}(g_{o,\tau}) \sim p_\Theta(\cdot\ |\ \tilde{I_o}, \tau, t_{\tilde{I_o}}(g_{o,\tau}))
\end{equation} 
This grasp orientation $R_{\tilde{I_o}}(g_{o,\tau})$ is represented as a 2D line, passing through the object at the task-oriented object affordance localization $t_{\tilde{I_o}}(g_{o,\tau})$, indicating the closing axis of the gripper in the object image $\tilde{I_o}$.
In practice, this is parametrized by 2 points, uniquely identifying the gripper's closing axis at pose $g_{o,\tau}$, in image space.
We then utilize the camera intrinsics $K \in \mathbb{R}^{3\times3}$ and extrinsics $T\in\mathbb{SE}(3)$ to lift the task-oriented object affordance localization as -
\begin{equation}
    t_{\mathbb R^3}(g_{o,\tau}) = f_{K,T}(t_{\tilde{I_o}}(g_{o,\tau}), D_o, r_o)
\end{equation} to 3D space, where $f_{K,T}$ is an abstract lifting function and $r_o$ is residual information regarding reconstructing $I_o$ from $\tilde{I_o}$.
We further lift the gripper's closing axis to 3D space as follows - 
\begin{equation}
    R_{\mathbb{SO}(3)}(g_{o,\tau}) = f_{K,T}(R_{\tilde{I_o}}(g_{o,\tau}), D_o, r_o)
\end{equation}

Note that $t_{\mathbb R^3}(g_{o,\tau})$ is a line and $R_{\mathbb{SO}(3)}(g_{o,\tau})$ represents a plane in 3D space, since their respective 2D counterparts in image space were a point and line in 2D space.
Finally, to obtain the vision-language prior for $g_{o,\tau}$, we define a multivariate distribution $\mathcal N(\mu^g_{o,\tau}, \Sigma^g_{o,\tau})$.
Our choice of $\mu^g_{o,\tau}$ is the grasp with translation being the point $q_t \in P_o$ closest to $t_{\mathbb R^3}(g_{o,\tau})$, and oriented so the plane of gripper closing is parallel to $R_{\mathbb{SO}(3)}(g_{o,\tau})$ and the gripper's approach axis is aligned with the camera $\Omega$'s view direction.
We further choose $\Sigma^g_{o,\tau}$ to heavily bias sampling around $q_t$ and in the plane $R_{\mathbb{SO}(3)}(g_{o,\tau})$.

\subsection{6DoF Grasp-Pose Refinement}
While the mean of the vision-language prior $\mu^g_{o,\tau}$ is a reasonable task-oriented 6 DoF grasp for the object $o$, it remains in the image plane of $I_o$.
As discussed earlier, this is a limitation which has the consequence of failing to consider a large proportion of feasible 6 DoF grasps that would satisfy our goal.
Additionally, small errors in depth estimation within object $o$'s depth map $D_o$, minor localization errors for $t_{\tilde{I_o}}(g_{o,\tau})$, and inference errors for $R_{\tilde{I_o}}(g_{o,\tau})$ can all compound to cause failures.
To address this, instead of using $\mu^g_{o,\tau}$ directly, TGrasp6D utilizes it as a prior $\mathcal N(\mu^g_{o,\tau}, \Sigma^g_{o,\tau})$ to inform a refinement procedure to produce final grasp output.
Our approach samples candidate grasps
\begin{equation}
    g^{\{N\}}_{o, \tau} \sim \mathcal N(\mu^g_{o,\tau}, \Sigma^g_{o,\tau})
\end{equation}
for $k = 1,\dots,N$, as an initial starting point.
Following this, we perform a number of optimization steps, independently for each candidate grasp $g^k_{o, \tau}$, using the following objective:

\begin{equation}
    \zeta^*_k = \argmin_{\zeta\in\mathbb R^6}[S(G_k(\zeta)) + \zeta^\top W\zeta]
    \label{eqn:opt_objective}
\end{equation}
for $k = 1,\dots,N$, where $G_k = g^k_{o, \tau}\exp(\hat\zeta)$, $\zeta \in \mathbb R^6$ is the optimization parameter, $\zeta^\wedge\in\mathfrak{se}(3)$ is a member of the Lie algebra of $\mathbb{SE}(3)$, and

\begin{equation}
\begin{aligned}
S(g) &= \frac{\lambda_c}{|P_o|} \sum_{q\in P_o}\max\{0,m_c-\Xi_g(q)\} \\
    &\quad - \frac{\lambda_\tau}{|P_o|} \sum_{q\in P_o}\nu_g(q)\rho_\tau(q) \\
\end{aligned}
\end{equation}
is the grasp scoring function which considers collision and task-consistency, with $m_c$ acting as the collision margin, $\lambda_c$ as collision coefficient, and $\lambda_\tau$ as the task-consistency coefficient.
Importantly, for the pose $g \in \mathbb{SE}(3)$, $\Xi_g: \mathbb R^3 \to \mathbb R$ is a signed distance field (SDF) which represents the gripper geometry and $\nu_g: \mathbb R^3 \to \{0, 1\}$ is an indicator function representing the sweep volume $\Psi(g) \subseteq \mathbb R^3$

\begin{equation}
    \nu_g(q) = \begin{cases}
    1 & \text{if } q \in \Psi(g)  \\
    0  & \text{if } q \not\in \Psi(g)
  \end{cases}
\end{equation}

and $\rho_\tau(q) = \mathcal N(\mu_\tau, \Sigma_\tau)$ is a multivariate function representing the ``task field.'' In practice, we use $\mu_\tau=q_t$ and treat $\Sigma_\tau$ as a hyperparameter. Finally, we select the best grasp as

\begin{equation}
    g^*_{o,\tau} = \argmin_{G_k(\zeta^*_k), k=1,\dots,N} [S(G_k(\zeta^*_k)) + (\zeta^*_k)^\top W\zeta^*_k]
    \label{eqn:opt_selection}
\end{equation}

This optimization step has, in particular, three important aspects: collision avoidance, task consistency, and pose delta-regularization.
The first term in the grasp scoring function $S$ penalizes collision between the gripper for a given grasp $g$ and object $o$'s point-cloud set $P_o$, so as to avoid grasps that intersect with the object geometry.
The second term in $S$ rewards the task-relevant object affordance point $\mu^g_{o,\tau}$ in the sweep volume $\nu$ of the gripper, encouraging the optimal grasp to preserve task-consistency.
And, lastly, the final term in the objective (Eqn.~\ref{eqn:opt_objective}) is a weighted regularization term on the optimization parameter $\zeta_k$, penalizing grasps further from the candidate grasp $g^k_{o, \tau}$, so as to not nullify the benefit of the vision-language-informed grasp-pose prior.

In practice, for the gripper geometry $\Xi$, TGrasp6D learns an SDF representation (using DeepSDF \cite{park2019deepsdf}) offline to enable fast optimization online during grasp-pose refinement, whereas $\nu$ is sufficiently estimated using a bounding box.

\section{Results}
\subsection{Setup} 
\textbf{Dataset and Metrics}. 
We evaluate on the TaskGrasp dataset~\cite{murali2021same}, which provides up to $25$ annotated task-oriented grasps per object--task pair.
We retain pairs with at least $20$ ground-truth grasp annotations, yielding $310$ evaluation instances.
A predicted grasp is considered successful if its gripper sweep volume attains at least $0.1$ IoU with any ground-truth grasp for the corresponding object--task pair, similarly to ~\cite{vuong2024language}. For all evaluations, we report grasp success rate (SR) and generation failure rate (GFR), where the latter measures the fraction of inputs for which no grasp is produced. For generated grasps classified as failures, we additionally report the orientation error (fOE) and position error (fPE) relative to the nearest ground-truth grasp.

\textbf{Hyperparameters}. We tune the hyperparameters for our approach using grid search over the principal parameters governing grasp-prior sampling and pose refinement. The final configuration is chosen based on performance on a small validation set and is used across all reported experiments. Unless otherwise stated, the same hyperparameter setting is applied to all object--task pairs.

\textbf{Hardware Configuration}. We conducted our experiments on a workstation equipped with an NVIDIA RTX 5080 GPU and a $20$-core Intel Core i7 CPU. Our baselines requiring re-training were trained locally on the RTX 5080, while author-provided pretrained weights were used for others. Real-world experiments were conducted using a Franka Research 3 (FR3) robotic manipulator equipped with the parallel-jaw gripper. All grasp synthesis and pose-refinement components were executed on the same workstation.

\subsection{Baseline Comparison}

\begin{table}[htbp]
    \centering\vspace{2.5mm}
    \begin{tabular}{lcccc}
    \toprule
        Method 
        & \shortstack{SR \\ $\uparrow$ (\%)} 
        & \shortstack{GFR \\ $\downarrow$ (\%)} 
        & \shortstack{fOE \\ $\downarrow$ (rad)} 
        & \shortstack{fPE \\ $\downarrow$ (cm)} \\
    \midrule
    
    GCNGrasp \cite{murali2021same} & 72.1
    & 3.21
    & 0.72~\scriptsize{$\pm0.35$}
    & 1.36~\scriptsize{$\pm0.82$} \\
    
    GraspGPT \cite{tang2023graspgpt} & 61.7
    & 20.90
    & 0.86~\scriptsize{$\pm0.38$}
    & 1.90~\scriptsize{$\pm1.75$} \\
    
    GraspMolmo \cite{deshpande2025graspmolmo} & 65.0
    & 19.94
    & 0.77~\scriptsize{$\pm0.39$}
    & 1.68~\scriptsize{$\pm1.41$} \\
    
    GraspCoT \cite{chu2025graspcot} & 7.5
    & 0.00
    & 0.93~\scriptsize{$\pm0.42$}
    & 4.45~\scriptsize{$\pm2.18$} \\
    
    F. Grasp \cite{tang2025foundationgrasp} & 73.7
    & 15.43
    & 0.79~\scriptsize{$\pm0.40$}
    & 1.46~\scriptsize{$\pm0.83$} \\
    
    \midrule
    
    TGrasp6D & \textbf{94.5}
    & 0.00
    & 0.77~\scriptsize{$\pm0.44$}
    & 1.91~\scriptsize{$\pm1.33$} \\

    TG-6D w/o PR & 89.4
    & 0.00
    & 0.73~\scriptsize{$\pm0.38$} 
    & 1.77~\scriptsize{$\pm1.12$} \\
    
    \bottomrule
    \end{tabular}
    \caption{\label{tab:metrics-ds}  Comparison with baseline methods. Note that fOE and fPE are reported only for grasps marked as failures. All baseline results use unmodified author-released code and, where available, author-provided pretrained weights; we do not alter any baseline's inference pipeline.
    }
    \vspace{-4mm}
\end{table}

We compare with state-of-the-art approaches for 6DoF task-oriented grasp synthesis on object--task pairs from the TaskGrasp~\cite{murali2021same} dataset.
We utilized author provided open-source implementations of GCNGrasp~\cite{murali2021same}, GraspGPT~\cite{tang2023graspgpt}, GraspMolmo~\cite{deshpande2025graspmolmo}, and GraspCoT~\cite{chu2025graspcot}, with FoundationGrasp~\cite{tang2025foundationgrasp} having been implemented based on details in the paper.
We re-trained GraspMolmo and FoundationGrasp on our system, while author provided model weights were used for GCNGrasp, GraspGPT, and GraspCoT.
Scoring based approaches are paired with the task-agnostic grasp synthesis methods detailed in their respective papers. To better understand the failure modes of our approach and the role of grasp-pose refinement in recovering from them, we also conduct an analysis of the different failure modes.

\renewcommand{\grenderwidth}{0.124\textwidth}
\newcommand{\gtaskwidth}{0.1\textwidth}

\newcommand{\methodlabel}[2]{%
    \makebox[\linewidth][c]{\small (#1) #2}%
}

\newcommand{\tasklabel}[1]{%
    \centering{#1}%
}

\begin{figure*}[htbp]
    \vspace{1mm}
    \centering

    {
    \setlength{\tabcolsep}{1pt}
    \renewcommand{\arraystretch}{1.0}

    \begin{tabular}{
        @{}
        >{\centering\arraybackslash}m{\gtaskwidth}
        *{7}{>{\centering\arraybackslash}m{\grenderwidth}}
        @{}
    }

        &
        \methodlabel{a}{GraspGCN} &
        \methodlabel{b}{GraspGPT} &
        \methodlabel{c}{F. Grasp} &
        \methodlabel{d}{GraspCoT} &
        \methodlabel{e}{GraspMolmo} &
        \methodlabel{f}{\textbf{Ours w/o PR}} &
        \methodlabel{g}{\textbf{Ours}}
        \\

        \tasklabel{\textbf{Measuring Cup}\\handover}
        &
        \includegraphics[width=\linewidth]{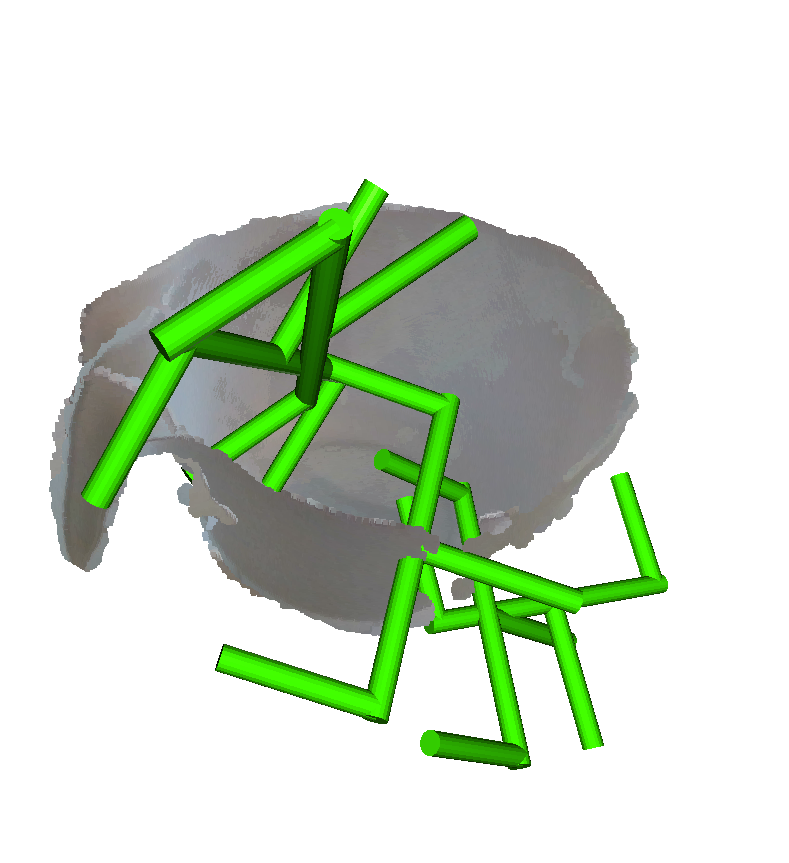}
        &
        \includegraphics[width=\linewidth]{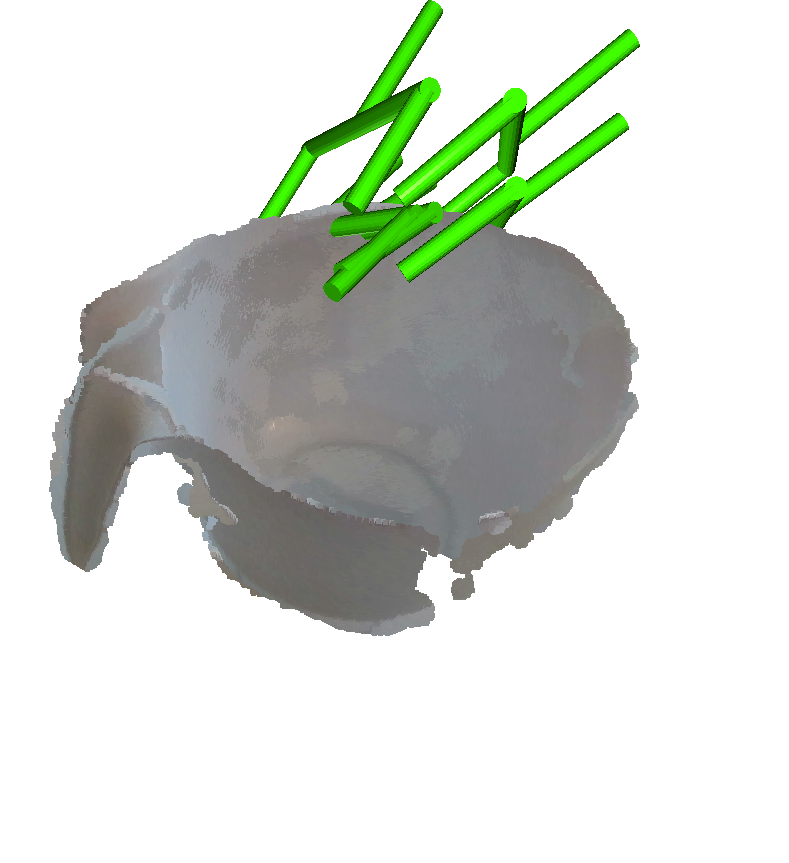}
        &
        \includegraphics[width=\linewidth]{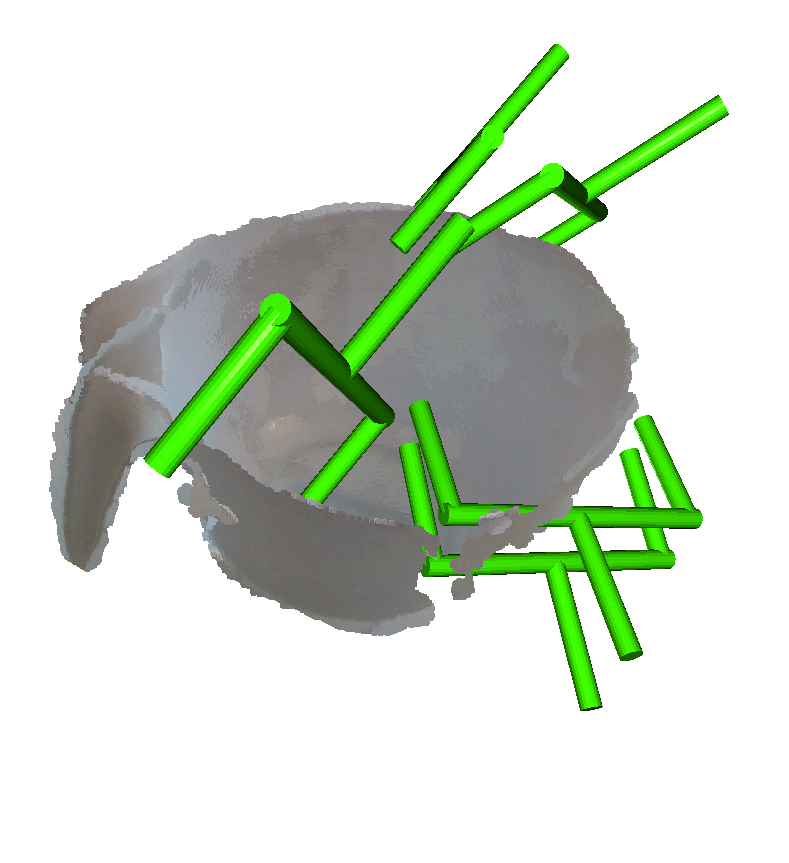}
        &
        \includegraphics[width=\linewidth]{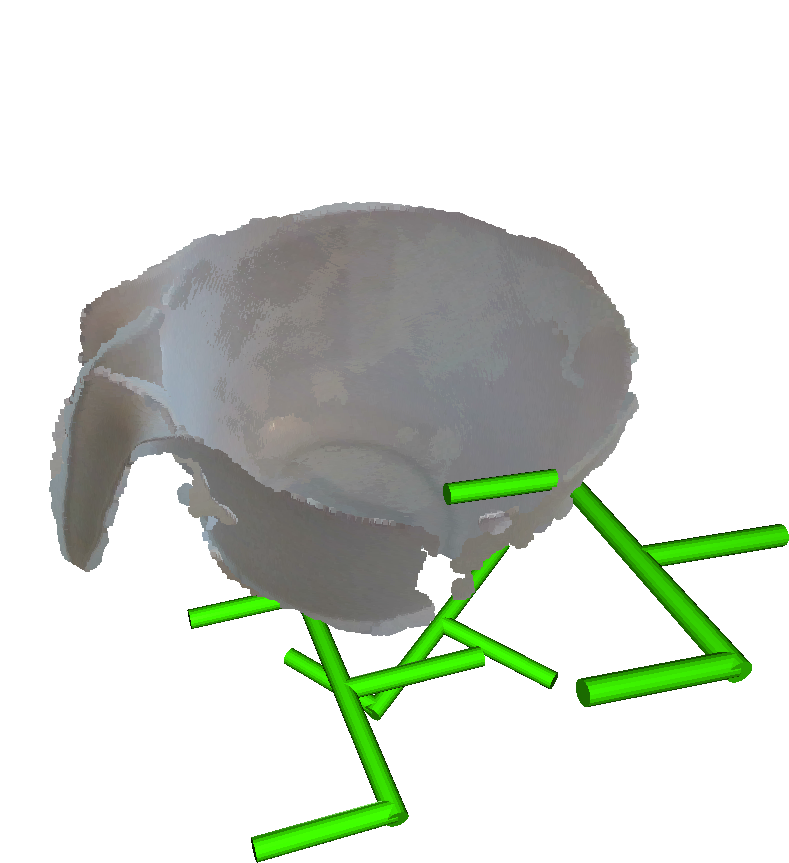}
        &
        \includegraphics[width=\linewidth]{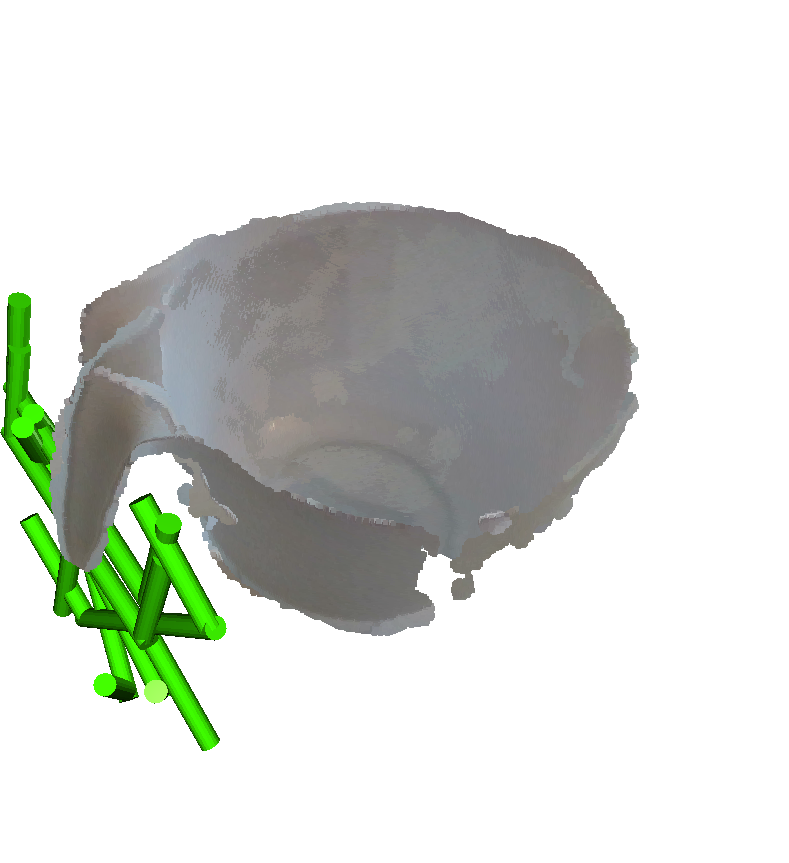}
        &
        \includegraphics[width=\linewidth]{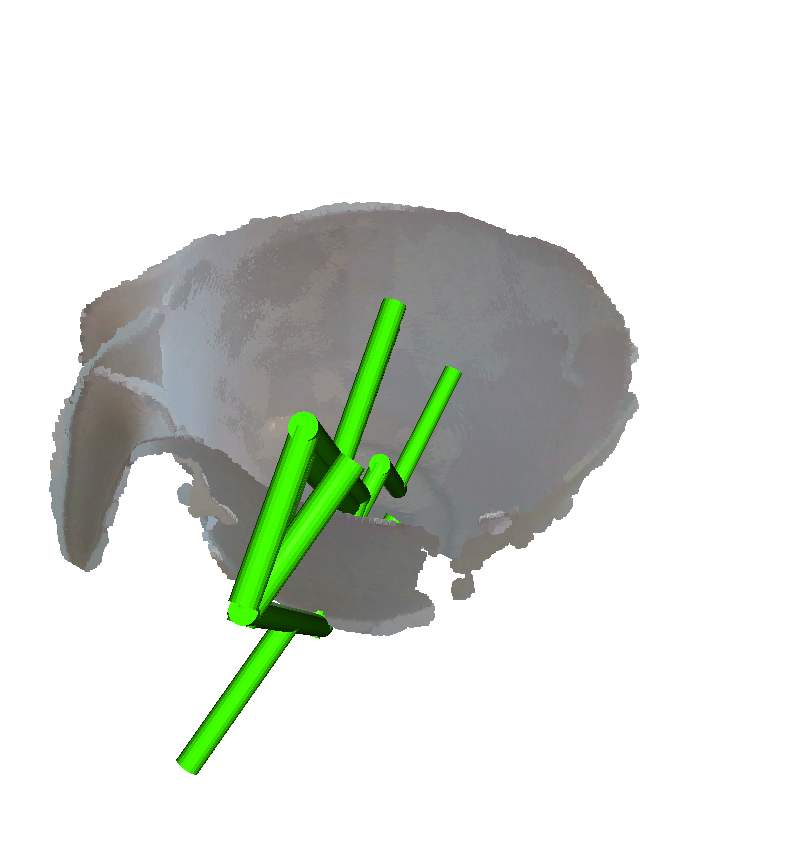}
        &
        \includegraphics[width=\linewidth]{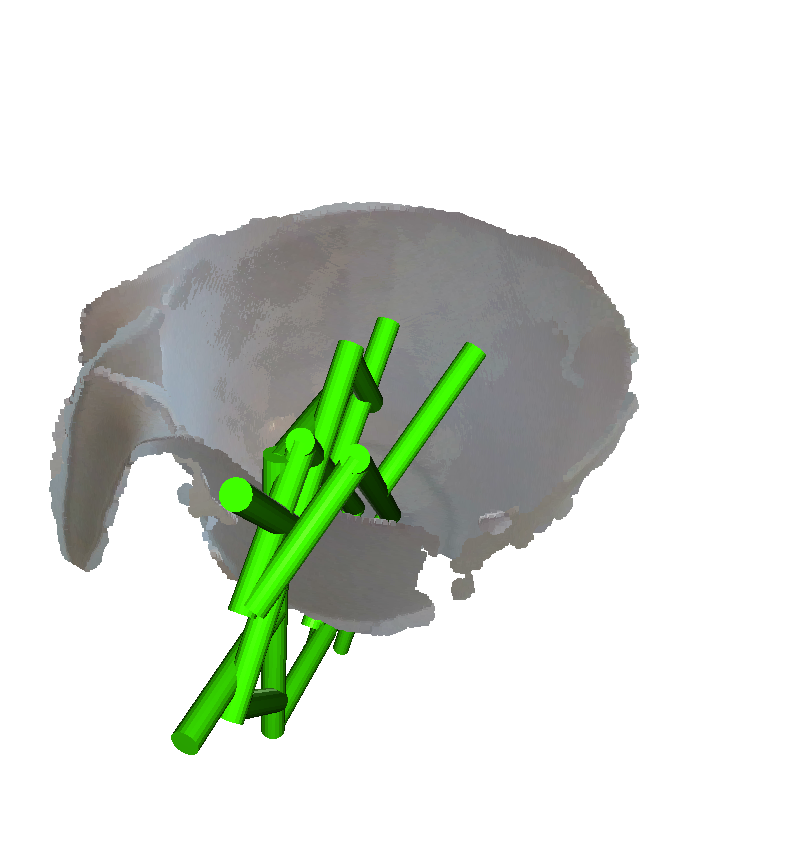}
        \\

        \tasklabel{\textbf{Paint Roller}\\handover}
        &
        \includegraphics[width=\linewidth]{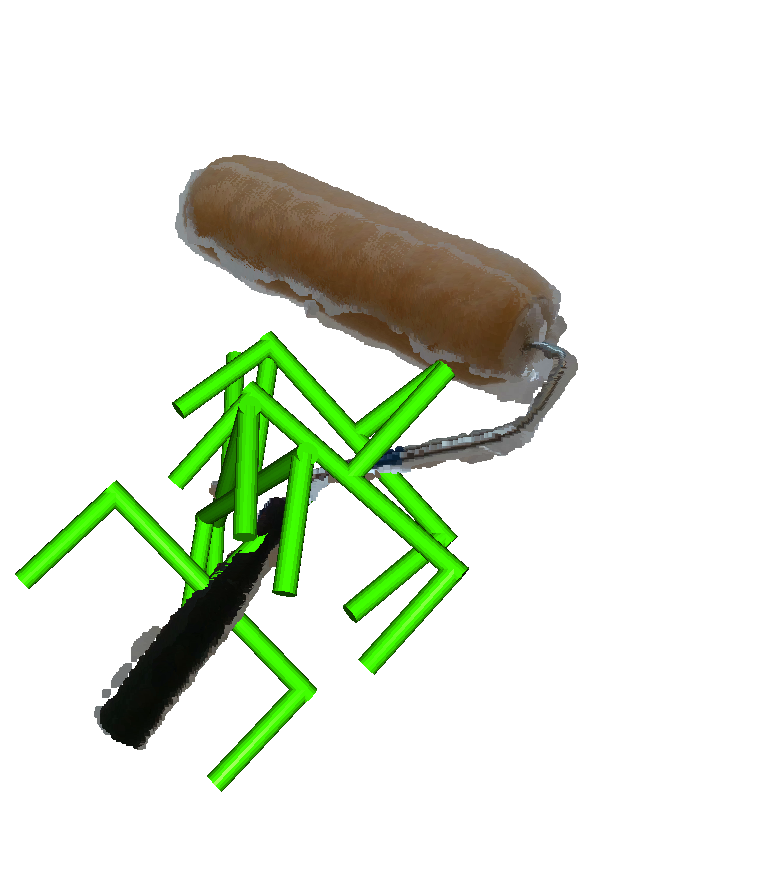}
        &
        \includegraphics[width=\linewidth]{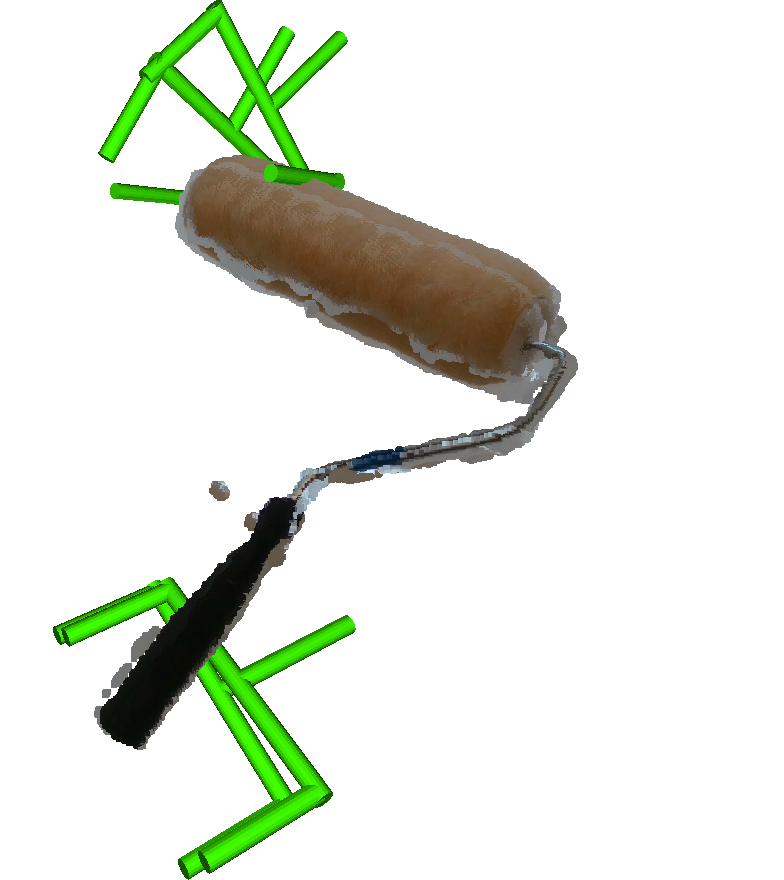}
        &
        \includegraphics[width=\linewidth]{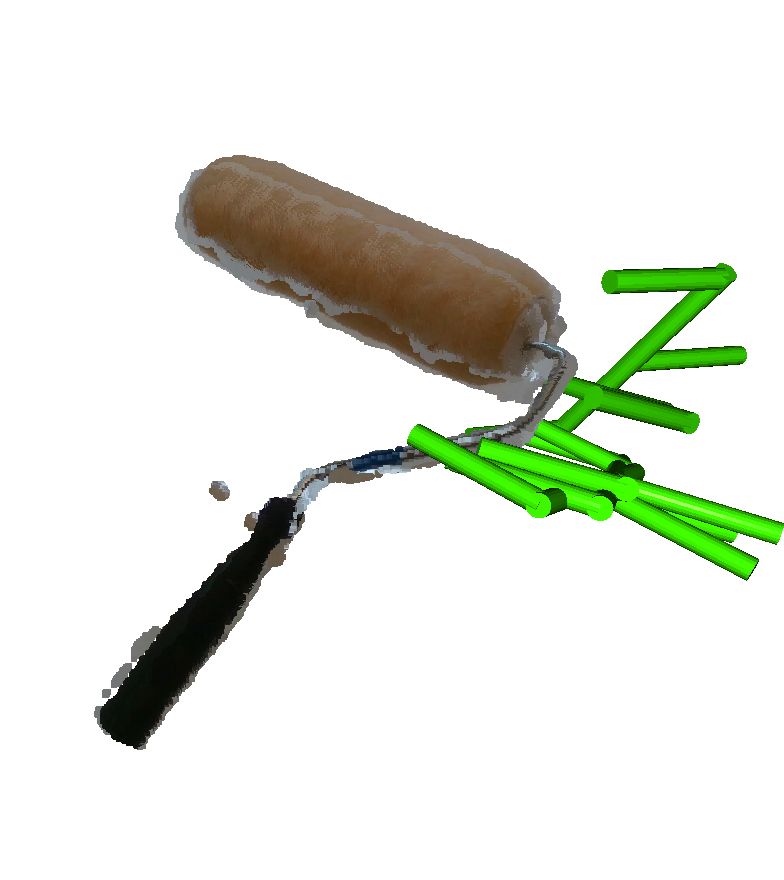}
        &
        \includegraphics[width=\linewidth]{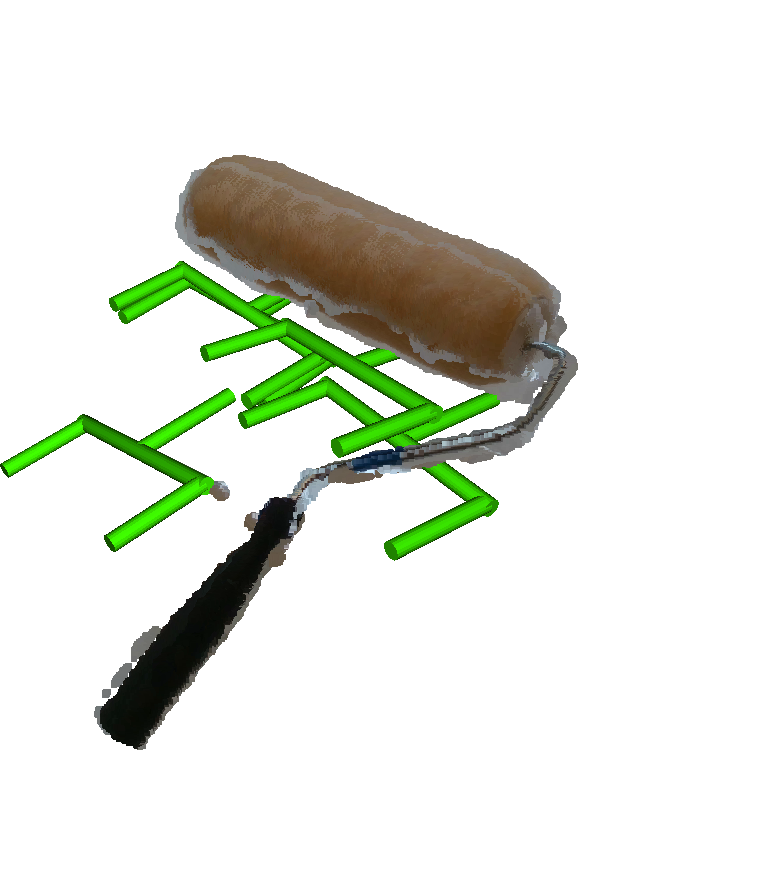}
        &
        \stackinset{c}{}{c}{}{%
            \shortstack{\textit{Failed}\\\textit{generation}}%
        }{%
            \phantom{%
                \includegraphics[width=\linewidth]{figs/gcomp/124_paint_roller/graspcot.png}%
            }%
        }
        &
        \includegraphics[width=\linewidth]{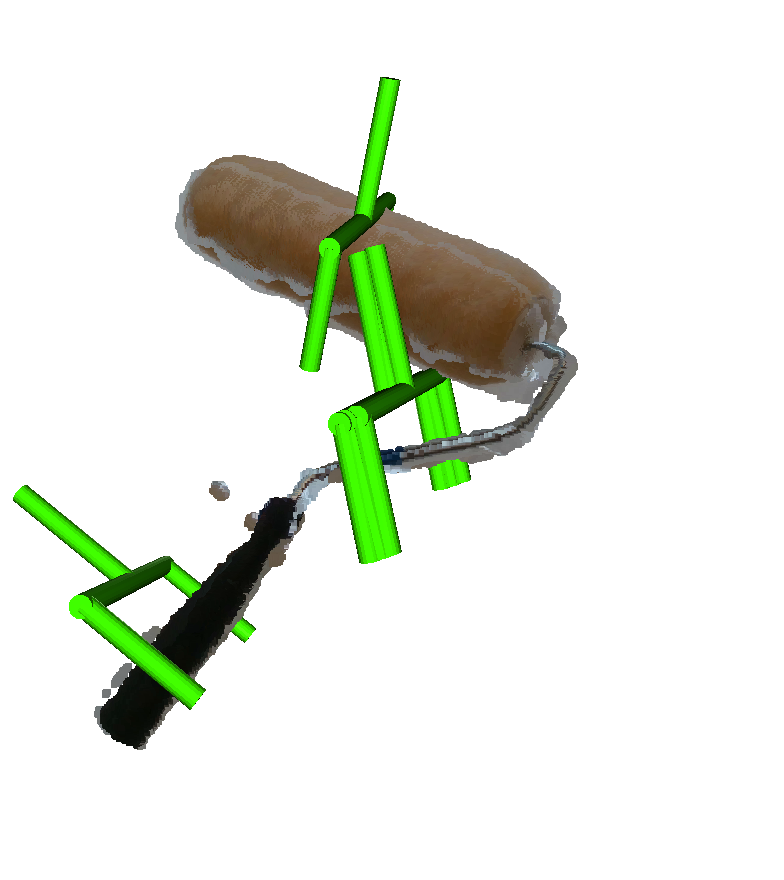}
        &
        \includegraphics[width=\linewidth]{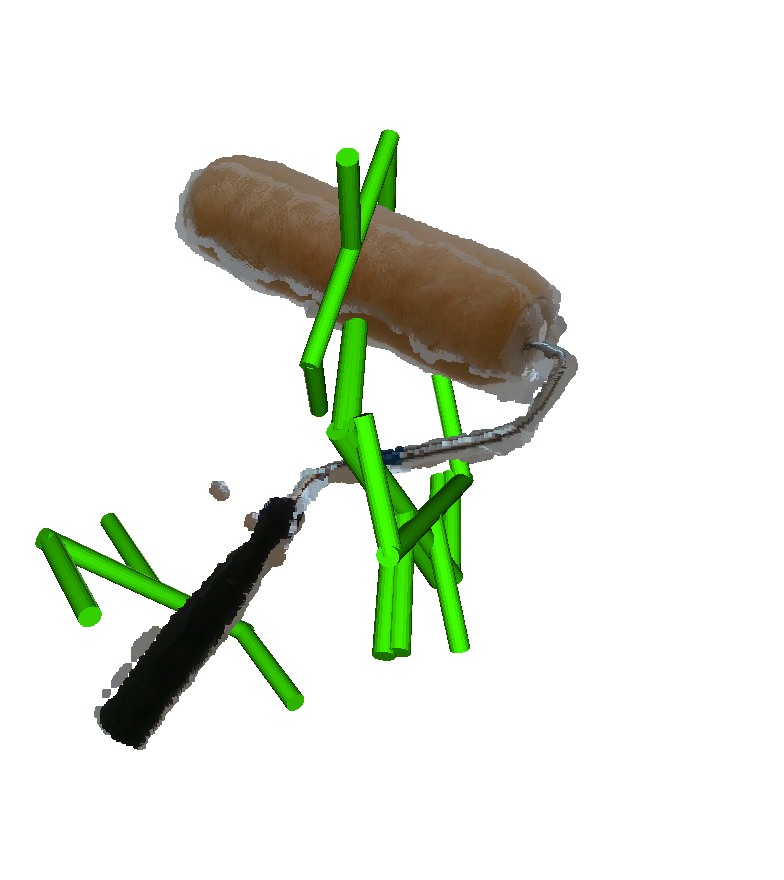}
        \\

        \tasklabel{\textbf{Masher}\\flatten}
        &
        \includegraphics[width=\linewidth]{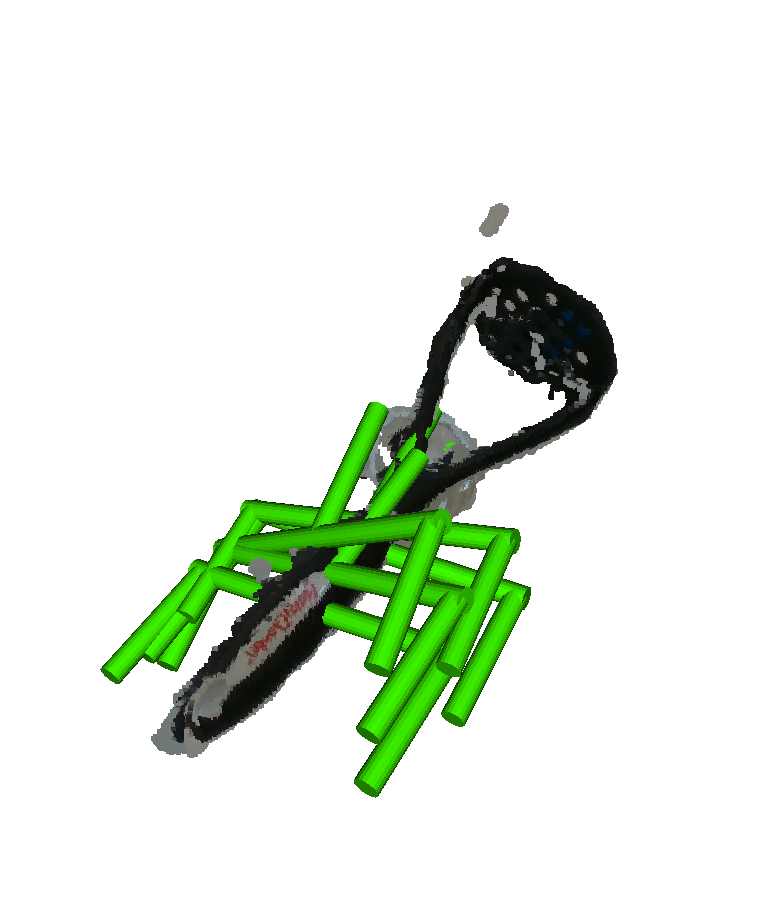}
        &
        \includegraphics[width=\linewidth]{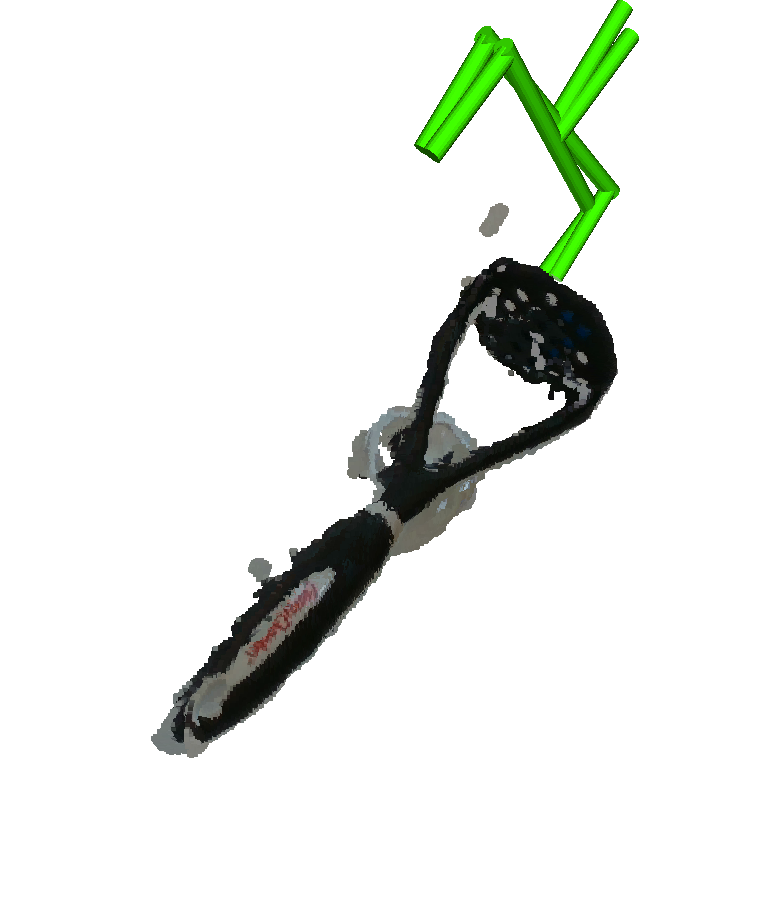}
        &
        \includegraphics[width=\linewidth]{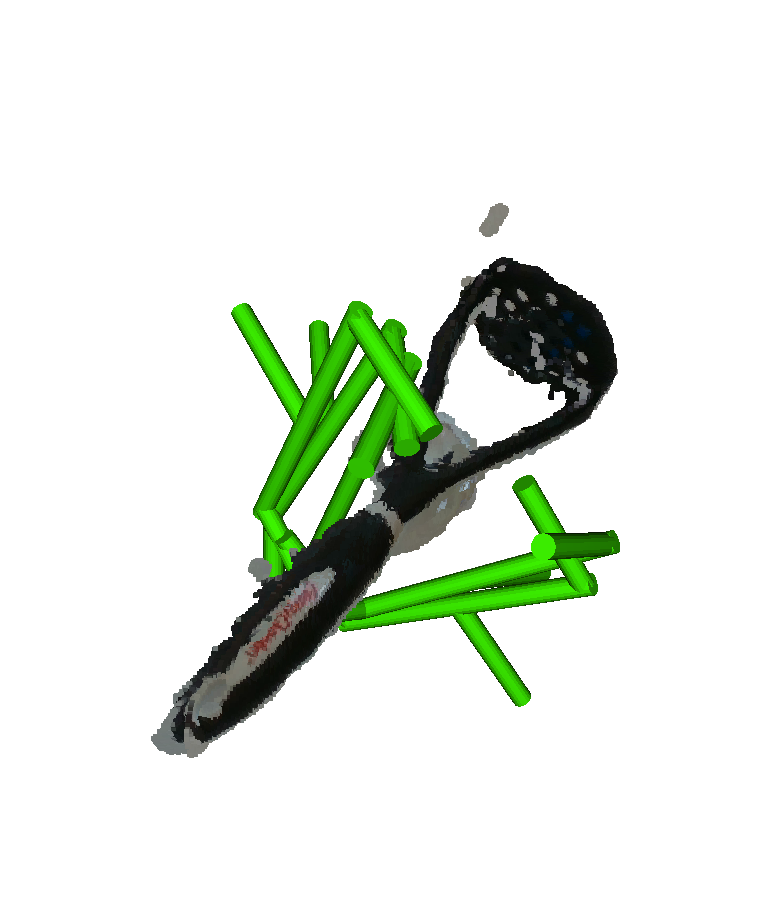}
        &
        \includegraphics[width=\linewidth]{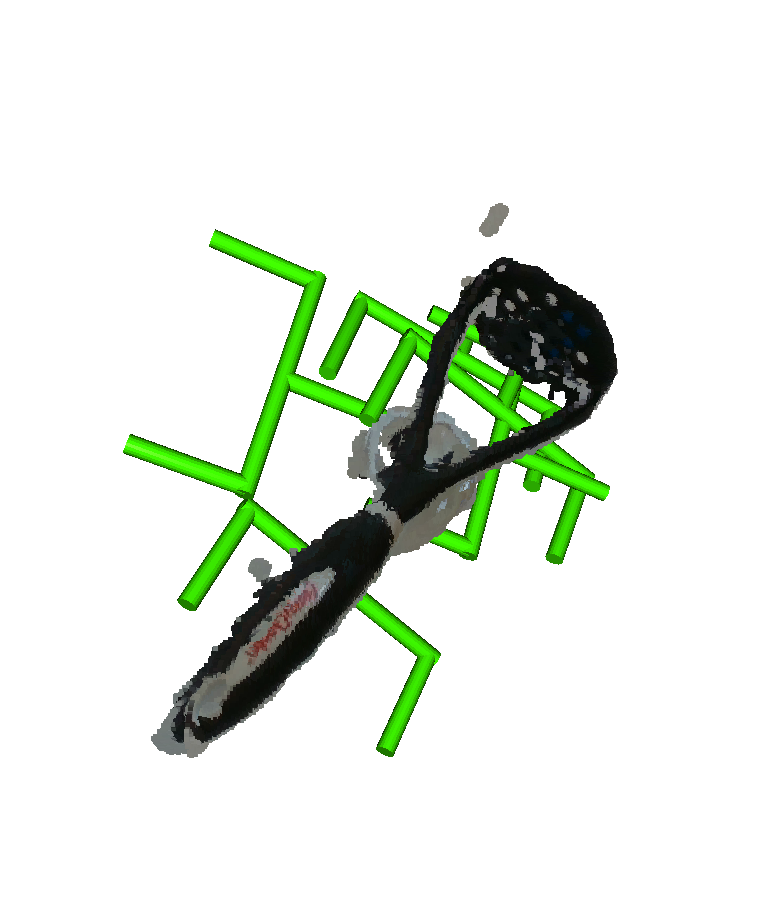}
        &
        \includegraphics[width=\linewidth]{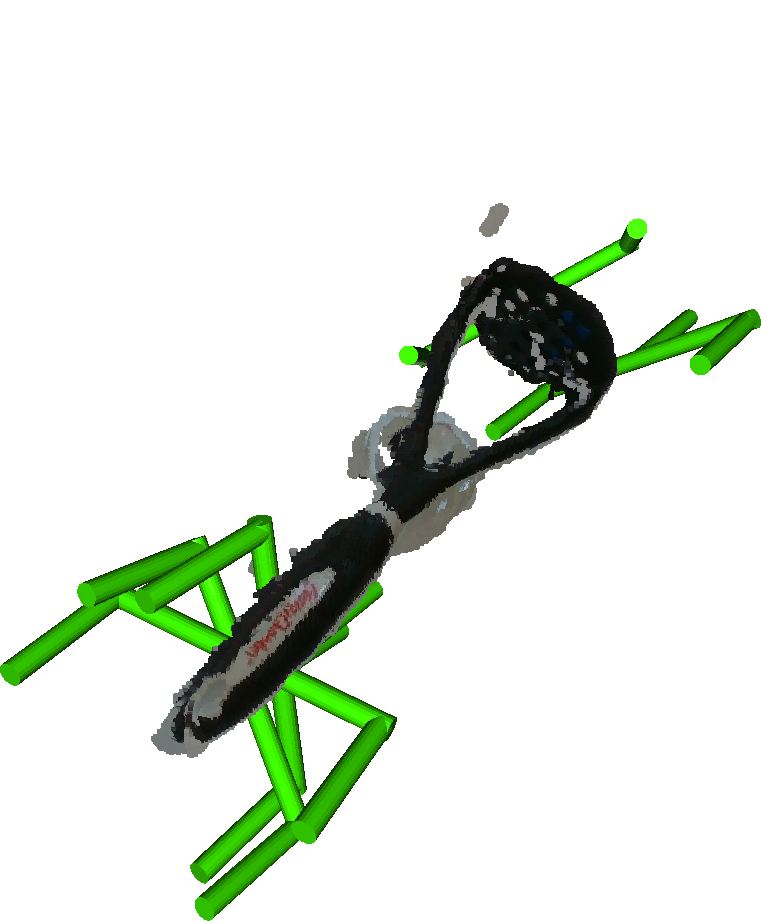}
        &
        \includegraphics[width=\linewidth]{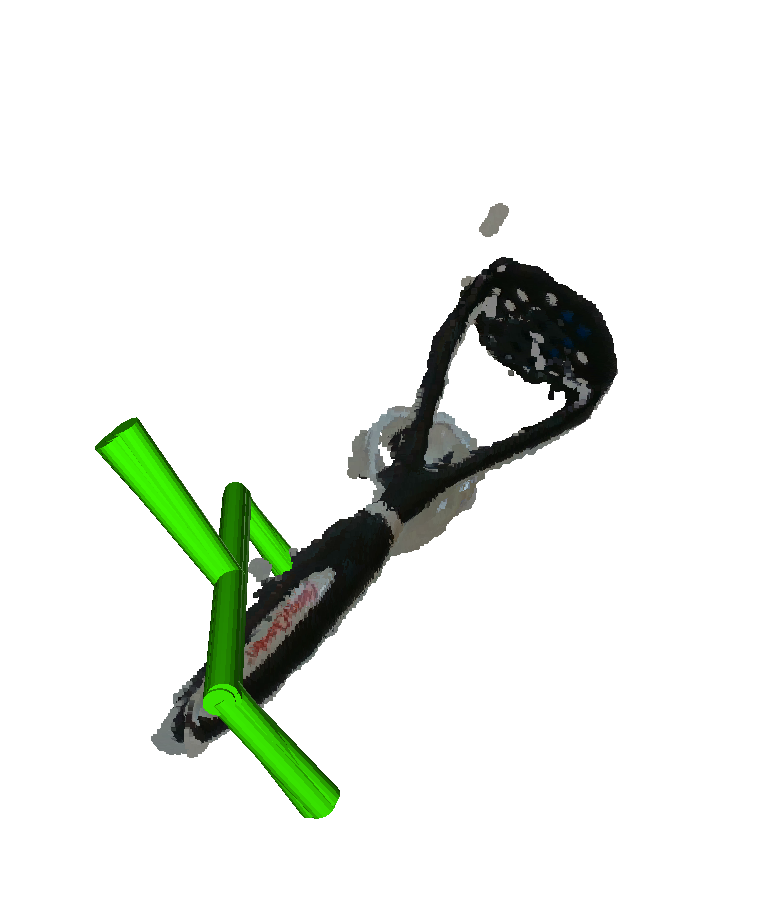}
        &
        \includegraphics[width=\linewidth]{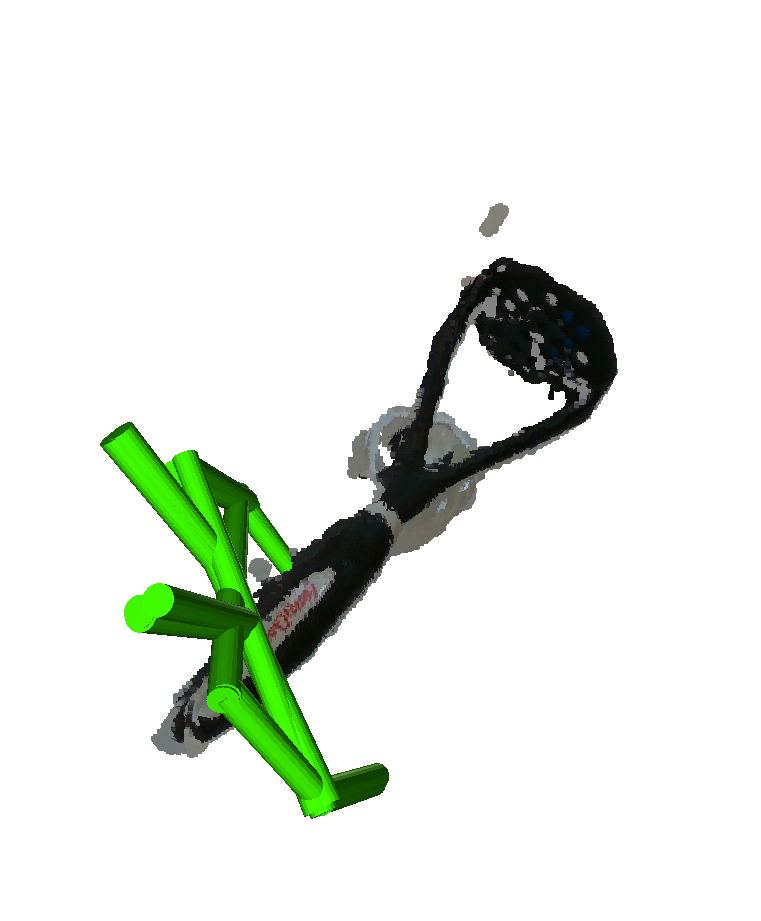}
        \\

        \tasklabel{\textbf{Spoon}\\scoop}
        &
        \includegraphics[width=\linewidth]{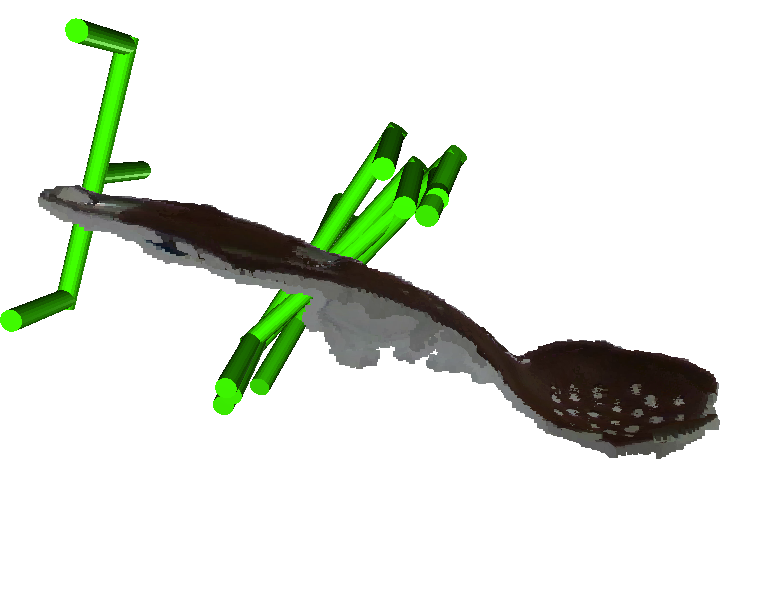}
        &
        \includegraphics[width=\linewidth]{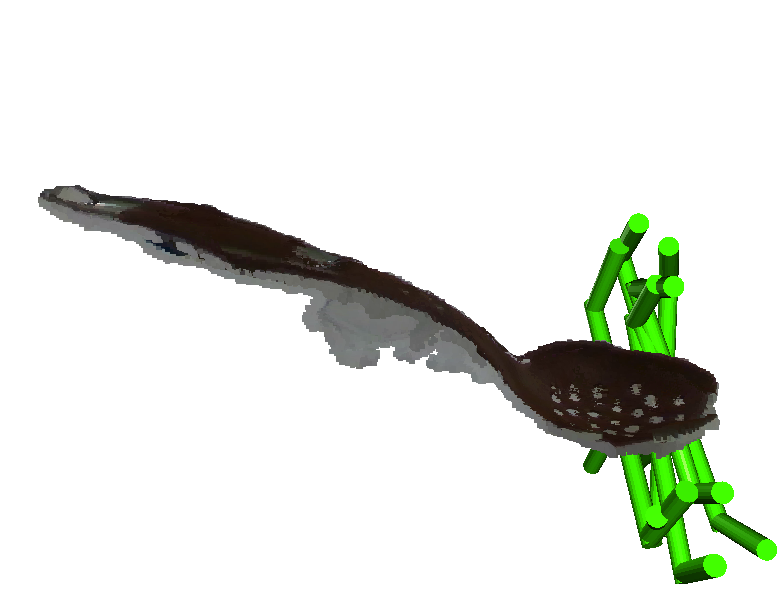}
        &
        \includegraphics[width=\linewidth]{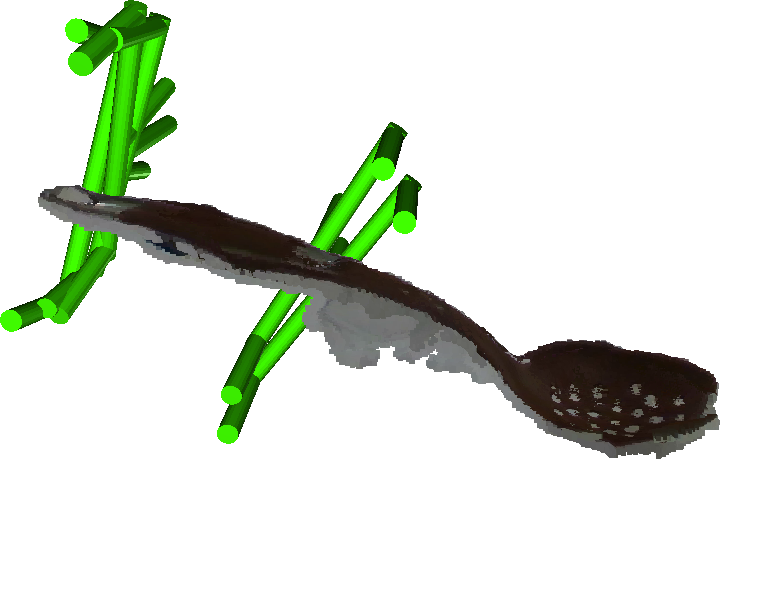}
        &
        \includegraphics[width=\linewidth]{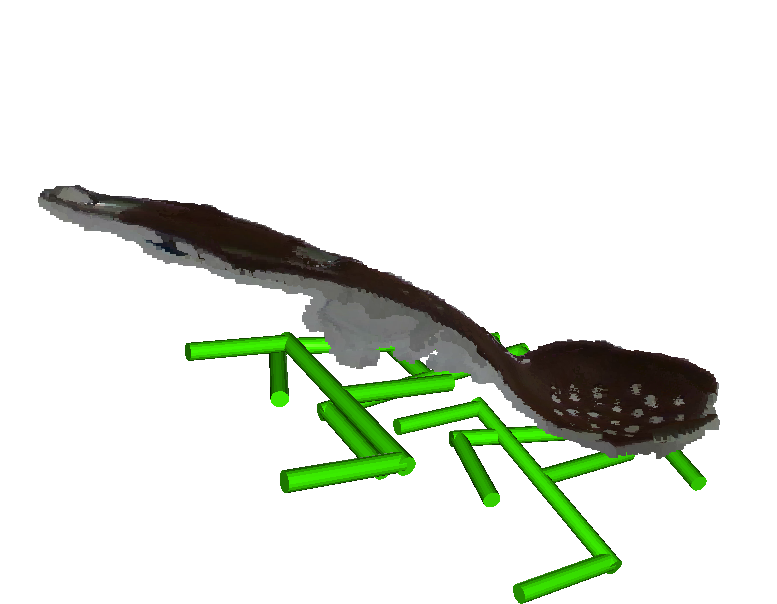}
        &
        \includegraphics[width=\linewidth]{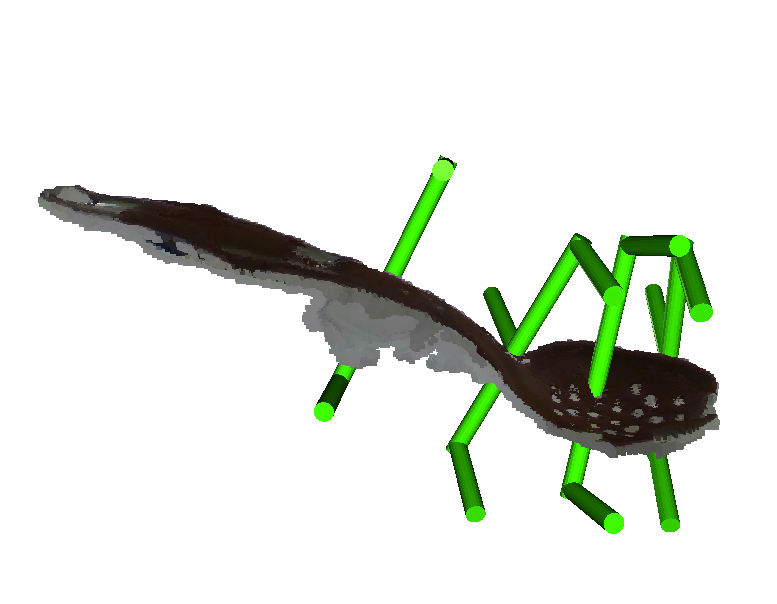}
        &
        \includegraphics[width=\linewidth]{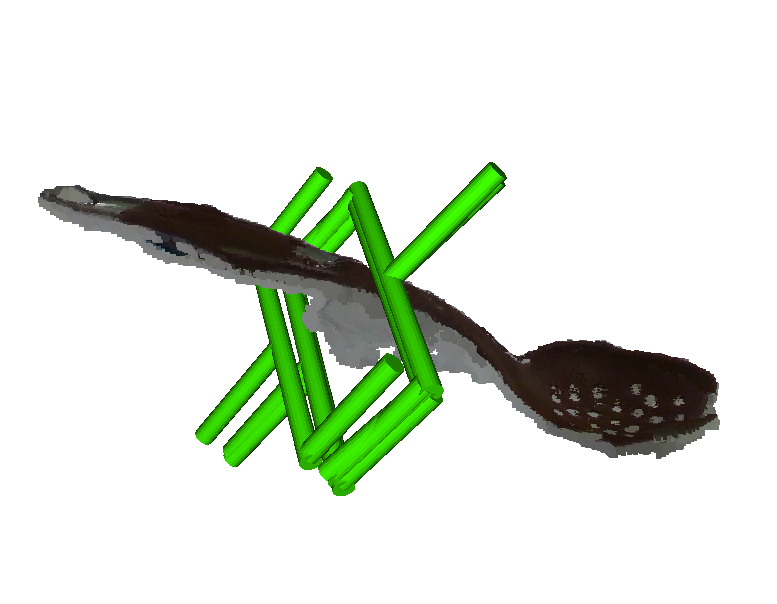}
        &
        \includegraphics[width=\linewidth]{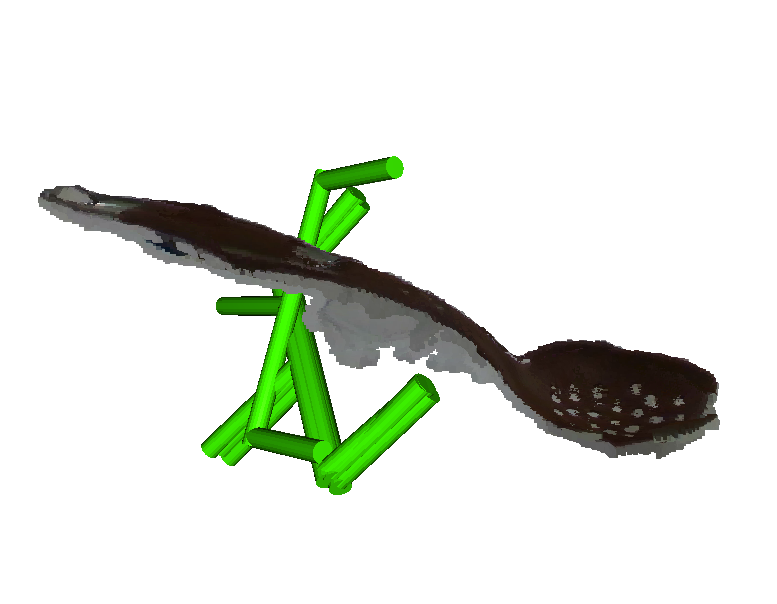}
        \\

    \end{tabular}
    }

    \caption{Qualitative comparison of different 6DoF TOG approaches on objects from the TaskGrasp~\cite{murali2021same} dataset. For each method, we show 5 task-oriented grasps per object. Compared to other approaches, our TGrasp6D produces more successful grasps across diverse object types and tasks. For cases like ``scoop[ing]'' with the ``spoon,'' most approaches generate feasible grasps. However, performance drops sharply for more challenging cases like doing a ``handover'' with the ``measuring cup.'' Of note, compared to the pure vision-language prior (f), grasp-pose refinement (g) appears to visually diversify grasps by generating feasible stable and task-consistent alternates. In some cases, like the ``measuring cup,'' it even recovers some failures despite noise and missing points due to occlusion in the object point-cloud.}
    \label{fig:gcomp}
    \vspace{-5mm}
\end{figure*}

Table~\ref{tab:metrics-ds} summarizes the quantitative baseline comparison. Our approach, TGrasp6D, achieves the highest success rate of $94.5\%$, exceeding the strongest baseline, FoundationGrasp at $73.7\%$. More importantly, the improvement is consistent across methods with substantially different supervision and inference strategies, suggesting that the gain is not solely attributable to a stronger semantic model. Instead, directly conditioning grasp synthesis on the task-relevant affordance appears to provide a stronger prior than generating task-agnostic candidates and imposing task semantics only during downstream selection.
With GPT-5.5 as the underlying VLM, our approach exhibits a 0\% generation failure rate, whereas several competing methods fail to return a grasp for a non-negligible fraction of inputs.
While our approach has higher fOE (0.77) and fPE (1.91) compared to other approaches, it is important to note that approaches with lower scores for these metrics are primarily training based and may overfit to grasp distributions seen during training. Also, note that these two metrics are reported only for grasps marked as failures. A higher value for these metrics may also indicate a larger spread/greater modality of possible generations that are not represented adequately in the human-annotated dataset.

The qualitative results in Fig.~\ref{fig:gcomp} further illustrate this difference across objects and tasks. While several methods produce suitable grasps for relatively unambiguous cases such as ``scoop'' with a spoon, performance becomes less consistent when the task requires selecting a specific functional region of the object. For example, during ``handover,'' the grasp must avoid obstructing the region intended for subsequent use. Our approach is more consistent in generating grasps satisfying such task-dependent constraints while remaining geometrically feasible.

Overall, our results support the central motivation of TGrasp6D: task semantics and geometric feasibility should be reasoned about jointly rather than sequentially, with a vision-language grasp prior grounding where to grasp and pose refinement ensuring the resulting grasp is stable and collision-free.

\subsection{Ablation Studies}
We conduct ablation studies along two principal dimensions of our approach: (i) the contribution of 6 DoF grasp-pose refinement, and (ii) the choice of vision-language model used for affordance reasoning, localization, and grasp-axis prediction. 
The former isolates whether geometric refinement meaningfully improves the task-informed prior by resolving collisions and correcting small localization or orientation errors, while the latter evaluates how sensitive the overall pipeline is to the semantic and spatial reasoning capabilities of the underlying VLM. 

\begin{figure}
    \vspace{1mm}
    \centering
    \includegraphics[width=\linewidth]{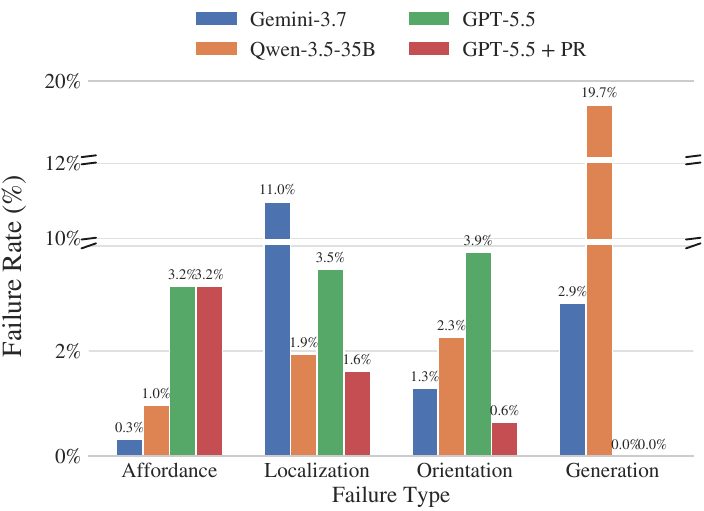}
    \caption{Analysis of failure modes, including the effect of grasp refinement for GPT-5.5. Note the y-axis breaks between 4\%-10\% and 12\%-19\% to accommodate Gemini-3.7 (11.0\%) and Qwen3.5-35B (19.7\%) without compressing the smaller values.}
    \label{fig:failure_pr}
    \vspace{-6mm}
\end{figure}

\textbf{Effect of Grasp-Pose Refinement}. 
We analyze the contribution of the 6DoF pose-refinement stage by evaluating TGrasp6D without refinement. 
As illustrated in Fig.~\ref{fig:failure_pr}, removing pose refinement decreases the success rate from $94.5\%$ to $89.4\%$, indicating that the vision-language prior alone provides a strong initialization, while geometric refinement recovers an additional $5.1$ percentage points. 
To better understand this improvement, we further decompose unsuccessful grasps according to their dominant failure mode: incorrect task-affordance selection, grasp-position error, or grasp-orientation error. 
Pose refinement substantially reduces position failures from $3.5\%$ to $1.6\%$ and orientation failures from $3.9\%$ to $0.6\%$, but not affordance failures which remain at $3.2\%$. 
This behavior is expected: pose refinement can correct geometric inaccuracies in the VLM-derived grasp prior, but cannot fully recover from an incorrectly inferred task-relevant affordance. 
These results suggest that refinement primarily improves geometric feasibility while preserving the semantic intent encoded by the initial prior. 
Qualitatively, refinement also produces a more diverse set of feasible grasps around the task-relevant region, as illustrated in Fig.~\ref{fig:gcomp}.

\textbf{VLM Ablation}.
We next study the sensitivity of our approach to the vision-language model used for affordance reasoning, localization, and grasp-axis prediction.
Results are reported in Table~\ref{tab:ablation_vlm}, with failure modes further decomposed in Fig.~\ref{fig:failure_pr}.
GPT~5.5 achieves the highest success rate of $89.35\%$, followed by Gemini~3.7 at $84.19\%$ and Qwen3.5-35B-4bit at $75.16\%$, yielding a $14.52$ percentage-point spread between the strongest and weakest VLM.
Importantly, the failure decomposition reveals substantially different behavior across models.
Gemini~3.7 exhibits strong affordance reasoning with only $0.3\%$ affordance failures, but suffers predominantly from localization errors ($11.0\%$), indicating that correct semantic reasoning does not necessarily translate to accurate spatial grounding.
In contrast, Qwen3.5-35B-4bit incurs relatively few semantic and geometric errors when it produces an output, but fails to generate a valid response for $19.7\%$ of inputs, making generation reliability its dominant bottleneck.
GPT~5.5 exhibits no generation failures and a more balanced error profile, with failures primarily arising from localization ($3.5\%$), orientation ($3.9\%$), and affordance reasoning ($3.2\%$).

The choice of VLM also substantially affects inference efficiency.
GPT~5.5 uses the fewest tokens across affordance identification, localization, and grasp-axis prediction, requiring on average $0.20\times10^3$, $0.34\times10^3$, and $0.61\times10^3$ tokens, respectively.
Overall, these results show that VLM selection affects not only grasp success and computational efficiency, but also the \emph{type} of errors propagated into the task-informed grasp prior, while TGrasp6D remains effective across substantially different VLM backbones.

\subsection{Real-Robot Deployment}

\begin{table}[t]
    \vspace{2mm}
    \centering
    \begin{tabular}{llccc}
    \toprule
    Object & Task & TGrasp6D (Ours) & GraspGPT & F. Grasp \\
    \midrule
    \multirow{2}{*}{Hammer}
        & pound & 10/10 & 2/10  &  5/10\\
        & handover & 10/10 & 6/10 &  5/10\\
    \midrule
    \multirow{2}{*}{Jug}
        & pour & 8/10 & 7/10 &  4/10\\
        & handover & 8/10 & 5/10 &  7/10\\
    \midrule
    \multirow{2}{*}{Pan}
        & cook & 10/10 & 6/10 &  3/10\\
        & handover & 10/10 & 3/10 &  5/10\\
    \midrule
    \multirow{2}{*}{Spray Bottle}
        & spray & 8/10 & 2/10 &  7/10\\
        & handover & 9/10 & 0/10 &  5/10\\
    \midrule
    \multirow{2}{*}{Sword}
        & stab & 10/10 & 0/10 &  5/10\\
        & handover & 10/10 & 10/10 &  6/10\\
    \midrule
    \multicolumn{2}{l}{Overall}
        & \textbf{93/100} & 41/100 &  52/100\\
    \bottomrule
    \end{tabular}
    \caption{ Real-robot task-oriented grasping success rate comparison.}
    \label{tab:metrics-rw}
    \vspace{-6mm}
\end{table}

Finally, we evaluate the zero-shot transfer of TGrasp6D to a Franka Research 3 robot using $5$ real-world objects and $10$ object--task combinations. 
Each combination is evaluated over $10$ trials, with object orientation varied across trials while keeping the orientation fixed across tasks for a given trial, yielding $100$ executions per method.
Results are summarized in Table~\ref{tab:metrics-rw}.
TGrasp6D succeeds in $93/100$ trials, compared with $41/100$ for GraspGPT and $52/100$ for FoundationGrasp.
Our approach achieves perfect success across both evaluated tasks for the hammer, pan, and sword, while maintaining $\geq80\%$ success for every object--task pair.

The dominant failures of TGrasp6D arise from imperfections in the observed geometry, particularly depth noise and self-occlusion.
For the ``jug,'' object, certain viewpoints occlude points near the desired grasp region, preventing reliable construction of a feasible grasp around the task-relevant affordance.
Similarly, for the ``spray'' task on the spray bottle, depth artifacts in some orientations cause otherwise suitable candidate poses to intersect the observed point cloud and therefore be rejected during collision-aware refinement.
The remaining failures originate from errors in the vision-language prior, including affordance localization or grasp-orientation errors, consistent with the failure modes identified in our ablation analysis.

Qualitative inspection of GraspGPT and FoundationGrasp reveals a different dominant failure mode.
Both methods depend on an upstream task-agnostic 6DoF grasp generator, and their performance is therefore constrained by the quality of the generated candidate set.
In several trials, the upstream generator produced grasps associated with noisy or incomplete point-cloud geometry, leaving the subsequent task-conditioned scoring stage to select among poor candidates.
We also observe semantic selection failures in which a geometrically valid grasp corresponds to an inappropriate task affordance; for example, grasping the sword by the blade for the ``stab'' task.
Together, these results suggest that TGrasp6D's real-world robustness stems from directly grounding candidate grasps in the task-relevant region, avoiding both the candidate-coverage limitations of an upstream task-agnostic generator and the semantic misselection that can persist even when a suitable candidate exists.

\begin{table}[t]
    \centering
    \vspace{2.5mm}
    \small
    \setlength{\tabcolsep}{4pt} %

    \begin{tabular}{@{}lcccc@{}}
    \toprule
    \multirow{2}{*}{VLM} &
    \multirow{2}{*}{SR (\%) $\uparrow$} &
    \multicolumn{3}{c}{\# Tokens Used $(\times10^3)$} \\
    \cmidrule(lr){3-5}
    & & Affordance & Localization & Grasp Axis \\
    \midrule

    GPT 5.5 &
    $89.68$ &
    $0.20$~\scriptsize{$\pm0.16$} &
    $0.34$~\scriptsize{$\pm0.34$} &
    $0.61$~\scriptsize{$\pm0.52$} \\

    Gemini 3.7 &
    $84.19$ &
    $0.78$~\scriptsize{$\pm0.74$} &
    $1.20$~\scriptsize{$\pm1.32$} &
    $1.80$~\scriptsize{$\pm1.28$} \\

    Qwen3.5-35B &
    $75.16$ &
    $2.12$~\scriptsize{$\pm1.69$} &
    $1.36$~\scriptsize{$\pm1.60$} &
    $5.09$~\scriptsize{$\pm2.08$} \\

    \bottomrule
    \end{tabular}

    \caption{Pre-pose refinement ablation using other vision-language models for affordance reasoning, localization, and grasp-prior generation.}
    \label{tab:ablation_vlm}
    \vspace{-6mm}
\end{table}

\section{Limitations and Future Work}

TGrasp6D currently relies on a single RGB-D observation, making grasp synthesis sensitive to depth noise, self-occlusion, and incomplete point clouds, which can cause the refinement stage to accept poses that collide with unobserved geometry. 
The method also depends on large VLMs, introducing inference latency and, for hosted models, API cost, while locally deployed alternatives can be computationally expensive. 
In addition, the current formulation localizes a single task-relevant region and therefore tends to concentrate grasp generation around one mode of the valid grasp space. 
Finally, the refinement objective does not explicitly model contact mechanics or closed-loop execution, motivating future work on multi-view perception, multiple affordance hypotheses, and feedback-driven grasping.

\section{Conclusion}

In this work, we presented TGrasp6D, a zero-shot framework for task-oriented 6DoF grasp synthesis from a single RGB-D observation and natural-language task instruction. 
By grounding task-relevant affordances with a vision-language model, lifting them into 3D grasp priors, and refining candidate poses for collision avoidance and task consistency, TGrasp6D removes the need for task-specific training data and task-agnostic grasp generators. 
On TaskGrasp, our method achieves a $94.5\%$ success rate, outperforming the strongest baseline by over $20$ percentage points, while real-robot experiments demonstrate strong zero-shot transfer across diverse object--task pairs. 
These results show that vision-language priors can effectively guide geometrically feasible, task-aware grasp synthesis without task-specific fine-tuning. 

\maketitle
\thispagestyle{empty}
\pagestyle{empty}

\renewcommand*{\bibfont}{\footnotesize}
\bibliographystyle{IEEEtranN}
\bibliography{IEEEabrv,root}

\end{document}